\documentclass[review, compress]{elsarticle}
\usepackage{array}
\usepackage{lineno,hyperref}
\usepackage{multirow}
\usepackage{amsmath,amsfonts}
\usepackage{booktabs}
\usepackage{siunitx}
\modulolinenumbers[5]

\journal{Pattern Recognition}

\begin{document}

\begin{frontmatter}

\title{DARTs: A Dual-Path Robust Framework for Anomaly Detection in High-Dimensional Multivariate Time Series}

\author[1]{Xuechun Liu}
\ead{luhan0420@stu.xjtu.edu.cn}

\author[1]{Heli Sun\corref{mycorrespondingauthor}}
\cortext[mycorrespondingauthor]{
 Corresponding author.}
\ead{hlsun@xjtu.edu.cn}

\author[1]{Xuecheng Wu}
\ead{wuxc3@stu.xjtu.edu.cn}

\author[1]{Ruichen Cao}
\ead{2127382011@stu.xjtu.edu.cn}

\author[1]{Yunyun Shi}
\ead{yunyunshi@stu.xjtu.edu.cn}

\author[2]{Dingkang Yang}
\ead{dkyang20@fudan.edu.cn}

\author[1]{Haoran Li}
\ead{lhr@xjtu.edu.cn}

\address[1]{Xi’an Jiaotong University, School of Computer Science and Technology, Xi’an, 710049, Shanxi, China}
\address[2]{Fudan University, College of Intelligent Robotics and Advanced Manufacturing, Shanghai, 200433, Shanghai, China}

\begin{abstract}
Multivariate time series anomaly detection (MTSAD) aims to accurately identify and localize complex abnormal patterns in the large‑scale industrial control systems. While existing approaches excel in recognizing the distinct patterns under the low‑dimensional scenarios, they often fail to robustly capture long-range spatiotemporal dependencies when learning representations from the high‑dimensional noisy time series. To address these limitations, we propose \textit{DARTs}, a robust long short-term dual‑path framework with window-aware spatiotemporal soft fusion mechanism, which can be primarily decomposed into three complementary components. Specifically, in the short-term path, we introduce a Multi‑View Sparse Graph Learner and a Diffusion Multi‑Relation Graph Unit that collaborate to adaptively capture hierarchical discriminative short‑term spatiotemporal patterns in the high-noise time series. While in the long-term path, we design a Multi‑Scale Spatiotemporal Graph Constructor to model salient long‑term dynamics within the high‑dimensional representation space. Finally, a window‑aware spatiotemporal soft‑fusion mechanism is introduced to filter the residual noise while seamlessly integrating anomalous patterns. Extensive qualitative and quantitative experimental results across mainstream datasets demonstrate the superiority and robustness of our proposed DARTs. A series of ablation studies are also conducted to explore the crucial design factors of our proposed components. Our code and model will be made publicly open soon.
\end{abstract}


\end{frontmatter}

\section{Introduction}
\label{sec:intro}
With the rapid evolution of information technologies and the proliferation of large-scale sensing, logging, and control infrastructures, multivariate time series anomaly detection (MTSAD) has become foundational for maintaining system integrity across domains such as industrial automation, financial surveillance, and intelligent manufacturing \cite{1}. Depending on the number of variables involved, MTSAD tasks fall into low-dimensional (\textit{i.e.}, $<100$ variables) and high-dimensional (\textit{i.e.}, $\geq 100$ variables) settings \cite{71}. Low-dimensional tasks are typically deployed in structurally simple systems such as embedded platforms, IoT devices, and home monitoring applications, where the interaction structure is sparse and stable. In contrast, high-dimensional MTSAD is indispensable in complex, large-scale systems including microservice infrastructures, algorithmic trading platforms, and cyber-physical energy grids, where cross-variable interactions are dense, multi-scale, and often nonstationary. Given its system-critical nature and increasing prevalence, high-dimensional MTSAD presents a particularly pressing and impactful research frontier.

Previous research widely relies on spatiotemporal integration for anomaly detection. Han and Woo~\cite{23} adopt a hybrid architecture with spatial self-attention and gated temporal convolutions to capture inter-sensor dependencies and causal temporal dynamics. Zhan et al.~\cite{25} employ stackable graph attention across feature and time dimensions to model multi-scale spatiotemporal correlations. Tang et al.~\cite{GRN} represent sensor interactions via dynamic graphs and preserves temporal continuity through residual connections. Zhan et al. Zheng et al.~\cite{CST-GL} integrate causal temporal convolutions with adaptive graph learning to jointly model temporal patterns and spatial dependencies. Pietroń et al.~\cite{AD-Nev} introduce a neural event-driven graph framework that encodes evolving spatiotemporal structures through dynamic graph construction and temporal convolutions. Cai et al.~\cite{FusionFormer} leverage a hierarchical transformer to decouple and fuse long-range temporal dependencies and inter-variable relationships via dual encoders across multiple datasets.

Although existing methods have achieved remarkable progress on the low-dimensional time series data, they still encounter substantial limitations when extended to high-dimensional contexts: \textbf{RQ1: How to effectively disentangle discriminative spatiotemporal signals under dimensional complexity?} (1) Existing approaches often fall short in decoupling hierarchical temporal dependencies, as window–context interference tends to obscure transient patterns with overarching temporal trends. (2) Furthermore, due to the constraints of high dimensionality and limited computational resources, existing methods struggle to comprehensively identify informative and multi-perspective inter-variable relations in a scalable and interpretable manner. \textbf{RQ2: How to enhance robustness against noise and redundancy in high-dimensional inputs?} Real-world high-dimensional time series are saturated with noise bursts, covariate shifts, and redundant channels, which obscure informative structures and undermine detection stability. Noise in time series data typically refers to unpredictable components that lack discernible patterns or structure~\cite{PR2}, and this noise further increases the complexity of high-dimensional data, hindering the identification of effective patterns and anomaly detection.

To tackle these challenges, we propose \textit{DARTs} (Dual-Path Robust Multi-Scale Time Series Anomaly Detection), a novel framework targetedly designed to address the spatiotemporal disentanglement and noise robustness in high-dimensional time series via a dual-path architecture and spatiotemporal soft fusion. Leveraging the dual-path design to enhance imformation concentration and cost efficiency, it decouples short-term signals and long-term trends to disentangle hierarchical temporal dependencies. The short-term path incorporates our proposed Multi-View Sparse Graph Learner and Diffusion Multi-Relation Graph Unit to isolate multi-view correlations and extract critical spatiotemporal structures from noise and redundant features. The long-term path introduces a Multi-Scale Spatiotemporal Graph Constructor to capture structured spatiotemporal dynamics in the noisy contexts. Subsequently, a window-aware spatiotemporal soft fusion mechanism is developed to dynamically filter dual-path outputs and emphasize salient signals. Extensive qualitative and quantitative experiments consistently demonstrate the superior interpretability, robustness, and efficiency of our model.

Beyond detecting anomalies in the high-dimensional multivariate time series, our introduced DARTs can generalize to modeling the intricate spatiotemporal dynamics, offering a feasible unified paradigm for forecasting and strategic decision-making in domains such as intelligent transportation and energy systems. In summary, the main contributions of this paper are three-fold: 
\begin{itemize}[leftmargin=*]

\item[$\bullet$] For high-dimensional MTSAD, we propose DARTs, a robust and scalable dual-path framework that separately handles transient patterns and overall temporal trends, demonstrating advantages in both computational efficiency and robustness. The framework aims to address the dual challenges of disentangling hierarchical spatiotemporal dependencies and enhancing resilience to dimensional noise.

\item[$\bullet$] The short-term path designs a Sparse Multi-View Graph Learner and a Diffusion Multi-Relation Unit to isolate inter-variable relations and extract hierarchical spatiotemporal structures in dimensional features. The long-term path introduces a Multi-Scale Graph Constructor to encode latent spatiotemporal trends from long noisy context sequences. A window-aware spatiotemporal soft fusion module integrates salient anomalies via resolution-aligned aggregation guided by instantaneous signals.

\item[$\bullet$] Extensive experiments across three datasets demonstrate that, under high-dimensional settings, our model outperforms existing state-of-the-art methods in terms of robustness and detection performance. Ablation studies further examine the contributions of each module. Interpretability experiments validate that the model is capable of identifying latent indirect anomalous variables in spatiotemporal anomaly propagation, even in the absence of a direct attack source.
 
\end{itemize}

The remainder of this paper is structured as follows. Sec. \ref{sec:Related Work} reviews and discusses related studies on spatiotemporal feature fusion and noise-robust strategies for multivariate time series (MTS) in both high-dimensional and low-dimens\\ional settings. Sec. \ref{sec:Problem} formally defines the problem of MTSAD. Sec. \ref{sec:Mthods} presents the architecture and training procedure of the proposed DARTs method. Sec. \ref{sec:exprs} describes the experimental setup, validation results, and comprehensive analyses. Finally, Sec. \ref{sec:Conclusions} discusses the advantages and limitations of our approach and outlines potential directions for future research.

\section{Related Work}
\label{sec:Related Work}

\subsection{MTSAD in Low and High Dimensions}

MTSAD spans a hierarchical spectrum from the low- to high-dimensional regimes. Low-dimensional settings typically exhibit sparse, near-stationary dependencies, enabling classical probabilistic models or lightweight deep networks to capture temporal dynamics at modest computational cost. High-dimensional scenarios, by contrast, often involve multiple loosely or tightly coupled subsystems with heterogeneous and evolving interdependencies that defy simple factorization. Two core challenges arise in such settings~\cite{73}:
(1) the difficulty of extracting anomalous spatiotemporal information stems from the exponential growth of the candidate channel-lag subspace with dimensionality and from asynchronous anomaly onsets across channels that obscure the true temporal footprint;
(2) signal-to-noise dilution, where irrelevant or weakly correlated channels inject spurious fluctuations that mask subtle anomalies, further complicating temporal localization.
These challenges necessitate expressive yet scalable representations capable of isolating relevant substructures, suppressing noise, and jointly modeling spatial and temporal anomaly signatures.

\subsection{Spatiotemporal Feature Integrations}

Existing approaches primarily focus on learning spatiotemporal representations for anomaly detection.~Zhan et al.~\cite{25} employ stacked spatial–temporal attention networks with multi-scale inputs to model hierarchical temporal patterns; however, its architectural complexity and sensor-specific preprocessing limit its scalability from low to highdimensional regimes.~Li et al.~\cite{14} fuse dilated convolutions with Transformer-based generators in an adversarial framework to enhance sequence fidelity, but incurs prohibitive training costs and lacks the capacity to capture hierarchical dependencies.~Yang et al.~\cite{27} introduce dual-path patch attention with contrastive learning, yet its channel-wise decomposition fails to model inter-variable correlations.~An et al.~\cite{28} combine dynamic graph attention with an Informer backbone, achieving improved accuracy on short sequences, but suffers from computational inefficiencies on longer and higher-dimensional inputs. Hypergraph-based models such as HgAD~\cite{29}, and spatiotemporal graph approaches like MAD-SGCN~\cite{30} and MTAD-GAT~\cite{38}, effectively encode higher-order or temporal dependencies, but exhibit limited scalability and considerable computational overhead, particularly under high-dimensional conditions. To address these limitations, we propose an efficient decoupled framework that separates short-term and long-term behaviors, thereby enhancing representational capacity and ensuring scalability. Additionally, we introduce a Multi-View Sparse Graph Learner for short-term processing, which adaptively infers hierarchical sparse topologies from instantaneous signals. For long-term processing, we propose a Multi-Scale Spatiotemporal Graph Constructor, which extracts multi-level spatiotemporal structure graphs from context memory in the presence of high-dimensional noise.

\subsection{Noise‑Robust Strategies}

Prior studies have explored rule-based methods, sparsity-regularized models, and hybrid generative-filtering paradigms for denoising in multivariate time series. For example,~Saravanan et al.~\cite{39} mitigate short-term noise through residual-based smoothing, but fails to address structural redundancy in high-dimensional contexts.~Zhou et al.~\cite{43} improve robustness via time–frequency augmentations and contrastive learning; however, its reliance on hand-crafted contrastive pairs limits scalability.~Wang et al.~\cite{8} handle data drift by decomposing and correcting spatiotemporal signals, though it suffers from high model complexity and a strong dependence on synchronized temporal embeddings.~Zhang et al.~\cite{45} constrain latent spaces using maximum mean discrepancy alignment, yet kernel selection remains nontrivial and poorly generalizes to high-dimensional, entangled data.~Han and Woo~\cite{23} enhance interpretability through sparse latent encodings, but its performance degrades significantly as dimensionality increases. In response to these limitations, we propose an efficient dual-path architecture that mitigates cross-scale window-context noise interference. Furthermore, we introduce a Multi-View Sparse Graph Learner in the short-term processing, which autonomously learns essential features and their relationships at varying hierarchical scales, thereby eliminating redundant feature structures in high-dimensional settings. In the long-term processing, we introduce a Multi-Scale Graph Constructor, which mitigates dimensional noise inherent in the long-term context by modeling multiresolution spatiotemporal dependencies. Lastly, a window-aware soft fusion module functions as the final filtering layer for irrelevant information within the embedding space, adaptively selecting anomalous information from context memory using instantaneous anomaly signals as the reference for retrieval.

\section{Problem Definition}
\label{sec:Problem}

Multivariate time series (MTS) consists of sequences of observations indexed by time, where each time point \( t \) is represented by a vector \( x_t \in \mathbb{R}^N \) of \( N \) distinct variables. The sequence has length \( T \), indicating the total number of time steps. 

In multivariate time series anomaly detection (MTSAD), the goal is to identify anomalies in the data. This task becomes challenging as the number \( N \) of variables increases, particularly for high-dimensional cases where \( N \geq 100 \). In such scenarios, detecting anomalies is complicated by intricate interactions among time-dependent variables, noise, and complex patterns that obscure anomalous signals.

Given a high-dimensional dataset, we split it into two subsets: \( X_{\mathrm{train}} \) containing normal data, and \( X_{\mathrm{test}} \) containing both normal and anomalous instances. The goal is to learn a robust detector function \( f: \mathbb{R}^N \to \{0,1\} \), where the input \( x_t \) at time step \( t \) is mapped to a binary label \( y_t \). Specifically, \( y_t = 1 \) indicates an anomaly, and \( y_t = 0 \) indicates normal data. Formally, the task is to solve:

\begin{equation}
y_t = f(x_t) \quad \text{where} \quad f(x_t) =
\begin{cases}
1 & \text{if } x_t \text{ is anomalous,} \\
0 & \text{if } x_t \text{ is normal.}
\end{cases}
\end{equation}

\section{Methodology}
\label{sec:Mthods}
\subsection{Overview}

\begin{figure}[t]
  \centering
  \includegraphics[width=\textwidth]{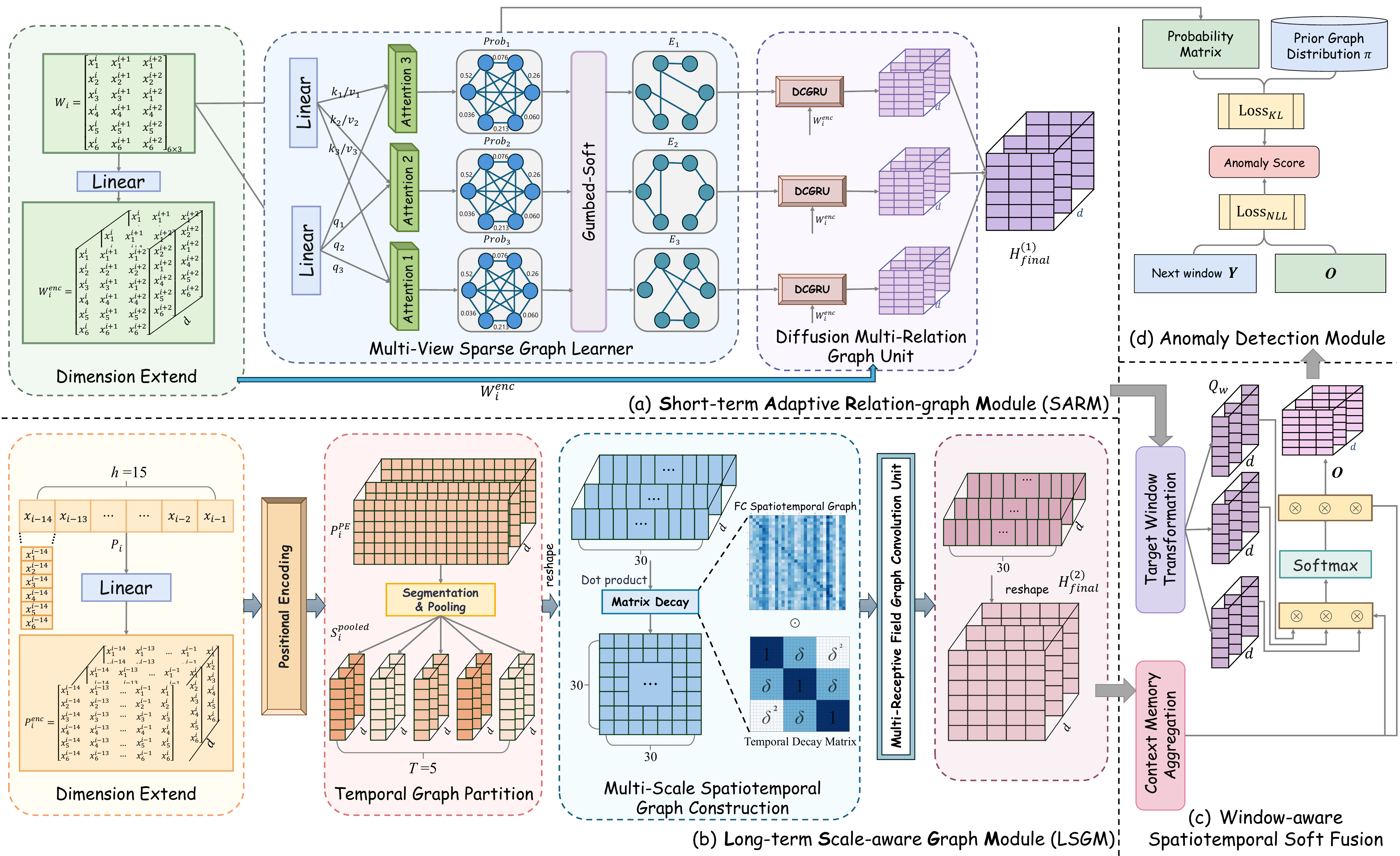}
  \caption{An overview of the proposed DARTs framework, comprising four main components: (a) Short-term Adaptive Relation-graph Module (SARM), (b) Long-term Scale-aware Graph Module (LSGM), (c) Window-aware Spatiotemporal Soft Fusion, and (d) Anomaly Detection Module.}
  \label{fig:framework}
\end{figure}

To address high-dimensional MTSAD tasks, we propose a dual-path decoupled architecture, DARTs, that offers both robust performance and scalability. As shown in Fig.~\ref{fig:framework}, the Multi-View Sparse Graph Learner and Diffusion Multi-Relation Graph Unit serve as the core components of the short-term path (SARM), jointly processing a short time window \( W_i = \{x_t\}_{t=i}^{i+w-1} \) to adaptively disentangle high-dimensional instantaneous signals into different hierarchical spaces and extract structured spatiotemporal patterns from redundant features and dimensional noise. Meanwhile, the long-term path (LSGM) is equipped with a Multi-Scale Spatiotemporal Graph Construction Module, encoding a broader, non-overlapping long-term history \( P_i = \{x_t\}_{t=i-h}^{i-1} \), where \( h \gg w \), to suppress spurious fluctuations introduced by irrelevant or weakly correlated channels in high-dimensional settings and capture significant delayed anomaly trends within the context. Furthermore, to enhance the integration capability and robustness of dual-path anomaly cues in high-dimensional spaces, DARTs introduces a window-aware spatiotemporal soft fusion mechanism that dynamically filters weakly correlated spatiotemporal information from context memory relative to instantaneous signals, while adaptively fusing anomaly dynamics from both paths.

\subsection{Short-term Adaptive Relation-graph Module}
\label{SARM}

Given an input window \( W_i \in \mathbb{R}^{N \times w} \) (where \( N \) is the number of variables and \( w \) is the number of time steps), we first apply a linear projection to obtain a feature-rich representation \( W_i^{\mathrm{enc}} \in \mathbb{R}^{N \times d \times w} \), where \( d \) denotes the latent dimension.

Firstly, we design a Multi-View Sparse Graph Learner to decouple hierarchical dynamic spatiotemporal dependencies in high-dimensional multivariate time series data, where the dynamic dependencies between variables are hidden variables that significantly affect the observable data~\cite{PR3}. Based on \( W_i^{\mathrm{enc}} \), we compute \( h \) independent attention heads, each representing a spatiotemporal observation perspective, yielding a set of attention matrices \( \text{probs} \in \mathbb{R}^{h \times N \times N} \). We then remove self-loops and apply a mask to obtain the attention mask matrix \( \text{probs}_{\text{masked}} \), which prunes redundant channel connections across different angles while preserving significant transient interaction patterns within the system. Its computation can be formally formulated as:

\begin{equation}
\text{probs}_{\text{masked}} = \text{probs} \odot M.
\end{equation}

To further enhance computational efficiency and reduce overfitting during training, we apply Gumbel-Softmax sampling to control the sparsity of the attention mask matrix, thereby reducing irrelevant or weakly correlated connections, \textit{i.e.},

\begin{equation}
E = \text{GumbelSoftmax}\left( \log(\text{probs}_{\text{masked}}), \tau \right),
\end{equation}
where \( \tau \in [0,1] \) represents a temperature parameter controlling the smoothness of the sampling. In our implementation, \( \tau \) is fixed at 0.5 throughout.

In each temporal-spatial domain, the sparse connectivity graph is constructed by selecting elements in the matrix \( E \) and assigning them a value of 1, where these elements correspond to the sampled edges. 

Additionally, to strengthen the robustness of the model, we assign prior knowledge with significant disparities in probabilities to each sparse graph during the training phase. This strategy ensures that graphs with higher probabilities focus on learning the most fundamental and simplest patterns in the early stages, while those with lower probabilities are encouraged to capture more complex multi-hop spatiotemporal dependencies. By constraining the connectivity between nodes, this approach effectively prevents the propagation and amplification of noise within the graph.

Ultimately, the Multi-View Sparse Graph Learner generates a multi-view sparse graph \( A \in \mathbb{R}^{\text{head} \times N \times N} \), which serves as evidence of the hierarchical structural relationships obtained by decoupling the instantaneous spatiotemporal signals in high-dimensional settings.

Subsequently, we propose a Diffusion Multi-Relation Graph Unit with parallel processing capabilities, designed to simulate anomalous hierarchical spatiotemporal propagation over multi-view sparse graphs. Initially, the temporal embeddings \( W_i^{\mathrm{enc}} \) and the multi-view sparse graph \( A \) are independently input into multiple Diffusion Convolutional Gated Recurrent Units (DCGRUs). Each DCGRU~\cite{46} performs multi-step propagation of node adjacency relationships within distinct spatiotemporal domains, while preserving the inherent temporal dependencies of the time series. The resulting encoded node representations capture asynchronous anomalous information induced by spatiotemporal diffusion following transient anomalies. 

Notably, in the Diffusion Multi-Relation Graph Unit, propagation is further limited by the sparse spatiotemporal topology learned by the Multi-View Sparse Graph Learner. Finally, we concatenate the DCGRUs outputs to obtain the stage-wise representation:

\begin{equation}
    \tilde{H}(h) = \text{Concat}\left(H_0(h), H_1(h), \dots, H_K(h) \right),
\end{equation}
where \( K \) denotes the number of diffusion steps on each sparse graph.

Finally, we fuse anomaly propagation features across spatiotemporal domains to obtain a unified transient anomaly diffusion embedding:

\begin{equation}
    H_{\text{final}}^{(1)} = \frac{1}{H} \sum_{head=1}^{H} \tilde{H}(head),
    \label{eq:h_final_1}
\end{equation}
where \( H_{\text{final}}^{(1)} \in \mathbb{R}^{N \times d \times w} \) represents the unified embedding, with \( N \) being the number of variables, \( d \) the latent dimension, and \( w \) the time steps of the short-term window.

\subsection{Long-term Scale-aware Graph Module}
\label{LSGM}

Given a historical sequence \(P_i\in\mathbb{R}^{N\times h}\) (with \(N\) high-dimensional variables and \(h\) time steps), we first lift \(P_i\) to a richer latent feature space to yield the encoded contextual trends
$P_i^{\mathrm{enc}}\in\mathbb{R}^{N\times d\times h}$, where \(d\) denotes the latent embedding dimension. To explicitly preserve the temporal order of the sequence and capture the long-range dependencies between variables, then we inject temporal information via Positional Encoding:
\begin{equation}
    P_i^{\text{PE}} = \text{PositionalEncoding}(P_i^{\text{enc}}) \in \mathbb{R}^{N \times d \times h}.
\end{equation}

Next, we partition the latent temporal representation into non-overlapping windows of length \(w\) and perform window‑wise average pooling to preserve the long-term patterns in the contextual memory:
\begin{equation}
S_i^{\text{pooled}} = \text{AvgPool}(\text{GraphPartition}( P_i^{\text{PE}}, w)) \in \mathbb{R}^{N \times d \times T},
\end{equation}
where \(T=\lfloor h/w\rfloor\) denotes the count of temporal windows.

To capture multi-granularity spatiotemporal dynamics across temporal windows under dimensional noise, we propose a Multi-Scale Spatiotemporal Graph Construction module, inspired by~Wang et al.~\cite{70}. This multi-scale module enables the model to capture rich spatiotemporal features and dynamic evolutionary interactions between different time scales. This is especially beneficial in handling the diverse dependencies within MTS~\cite{PR6}.

Specifically, we begin by projecting the window-pooled representations \( S_i^{\mathrm{pooled}} \) into a latent space \( S' = W\,S_i^{\mathrm{pooled}} \), where \( W \in \mathbb{R}^{d \times d} \). We then construct a fully connected (FC) spatiotemporal affinity graph \( A = S' {S'}^{\top} \) by computing pairwise dot-product similarities. This operation helps establish signal connections between channels at distinct time steps, while the graph elements encode multi-scale dependencies among temporal moments in the historical context. Subsequently, to preserve essential overarching trends, enhance the robustness of anomaly detection, and mitigate the effects of self-loops and redundancy, we apply a diagonal masking operation followed by row-wise softmax normalization to the spatiotemporal affinity graph, \textit{i.e.},
\begin{equation}
    A_{ij} = \text{softmax}(\text{LeakyReLU}(A_{ij} - \infty \cdot \kappa_{ij})), \quad \kappa_{ij} = 
    \begin{cases} 
    1, & \text{if } i = j, \\
    0, & \text{if } i \neq j,
    \end{cases}
\end{equation}
where \(\kappa_{ij}\) denotes the Kronecker delta and we use a small numerical constant \(10^{-6}\) for stability.

Finally, to simulate the propagation dynamics of anomalous signals in the context, which progressively attenuate over time, we employ a bidirectional temporal decay scheme to further refine long-range spatiotemporal dependencies:
\begin{equation}
D_{ij}=\delta^{\,|i-j|}, \quad \delta\in[0,1], \qquad A \;\leftarrow\; A \odot D,
\end{equation}
where \(\odot\) denotes the Hadamard product. Distance-based decay mitigates spurious long-range spatiotemporal correlations, thereby improving the robustness of the spatiotemporal affinity graph to noise inherent in long sequences.

In the subsequent stage, to further explore the propagation effects of the window-aggregated latent representation \( S_i^{\mathrm{pooled}} \) on the multi-granular spatiotemporal affinity graph \( A \), we simulate the diffusion process of contextual anomaly dynamics within the global trend context. We perform feature-wise normalization on the latent representation to obtain \( S_i^{\mathrm{normed}} \in \mathbb{R}^{N \times d \times T} \). Then, we apply a multi-receptive-field graph convolution unit~\cite{49} to encode the propagation of anomaly signals within the contextual memory:
\begin{equation}
H^{(r)} \;=\; \Theta^{(r)}\bigl(A^{(r)} S_i^{\mathrm{normed}}\bigr),
\end{equation}
where \(r\) indexes the receptive‑field scale and \(\Theta^{(r)}\) denotes the learnable
parameters at scale \(r\).

Afterwards, we flexibly fuse multi-scale encodings and apply node-wise normalization, yielding noise-resilient, scale-aware representations. Specifically, let
$\mathcal{S}=\{1,\dots,R\}$ denote the set of scale indices, where $R$ represents the maximum receptive-field size. We fuse $\{H^{(r)}\}_{r\in\mathcal{S}}$ via
\begin{equation}
\begin{split}
H_{\mathrm{fused}} &= \mathcal{F}_{\tau}\!\left(\{H^{(r)}\}_{r=1}^{R}\right), \quad \tau \in \{\text{concat},\, \text{sum}\}, \\
\text{followed by normalization:} \quad & \frac{H_{\mathrm{fused}} - \text{mean}}{\text{var}} + \epsilon \cdot \alpha + \beta.
\end{split}
\end{equation}

In the end, $H_{\mathrm{final}}^{(2)}$ is computed by normalizing the fused representation as shown below.
\begin{equation}
\label{eq:h_final_2}
H_{\mathrm{final}}^{(2)}
\;=\;
\frac{H_{\mathrm{fused}}-\mathrm{mean}(H_{\mathrm{fused}})}{\mathrm{var}(H_{\mathrm{fused}})}
\;+\; \varepsilon\cdot\alpha \;+\; \beta,
\end{equation}
where \( \alpha \) and \( \beta \) are learnable scale and shift parameters, and \( \varepsilon \) denotes a small constant for numerical stability. The resulting vector \( H_{\mathrm{final}}^{(2)} \in \mathbb{R}^{N \times d \times T} \) represents the final normalized embedding, where \( N \) is the number of variables, \( d \) is the latent dimension, and \( T \) denotes the time steps of the long-term sequences.

\subsection{Window-aware Spatiotemporal Soft Fusion}
\label{WASSF}
To enhance the effectiveness of the feature fusion process and improve both the detection performance and robustness of the model, we introduce a window-aware soft fusion module. This module aligns the long-term context embedding \( H_{\mathrm{final}}^{(2)} \) (Eq.~\ref{eq:h_final_2}) with the short-term instantaneous window \( H_{\mathrm{final}}^{(1)} \) (Eq.~\ref{eq:h_final_1}), dynamically selecting the most relevant anomaly signals and filtering out weakly correlated information. This fusion strategy strengthens the model's ability to detect anomalies more accurately.

The alignment of the two embeddings is achieved through an initial normalization and reshaping process, ensuring consistency within the vector space. The resulting aligned vectors are represented as \( T1 \in \mathbb{R}^{w \times N \times d} \) and \( T2 \in \mathbb{R}^{T \times N \times d} \), where \( w \) and \( T \) denote the time steps of the short-term windows and long-term sequences, respectively.

In real-world scenarios, anomaly detection is heavily influenced by the relationship between instantaneous patterns and the broader context. From a temporal perspective, long-term context provides essential historical background and clues that help uncover the underlying causal mechanisms behind anomalies. Causality-based feature selection methods play a vital role in this process, offering deeper insights into system behavior while enhancing interpretability~\cite{PR4}. Conversely, the short time span of the instantaneous window makes it more tightly linked to anomaly events, enabling it to directly capture the immediate characteristics of the anomalies. Hence, we prioritize short-term window embeddings as the main signal for anomaly detection, supplemented by long-term context embeddings to enrich and support the main signal.

To perform the feature fusion, we project the target window \( T1_w \in \mathbb{R}^{N \times d} \) into a query matrix \( Q_w \), while the context memory \( T2 \) serves as the key \( K \) and value \( V \). For each query window \( w \), we execute cross-attention from \( Q_w \) to the shared keys and values of the context, yielding window-level fused features. This process is expressed as:

\begin{equation}
    Z_w = \text{Softmax}\left(\frac{Q_w K^\top}{\sqrt{d_k}}\right)V,
\end{equation}
where $d_k$ refers to the dimensionality of query vectors.

Finally, we concatenate the window-level fused features, augmented with supplementary contextual information, and utilize a linear layer as the prediction mechanism to project the concatenated fused vector into a dimensional space consistent with the input of the short-term path (SARM), thereby yielding the predicted value for the subsequent time-step instantaneous window, \textit{i.e.},
\begin{equation}
    O = \text{Linear}_{\text{out}} \left( \text{Concat}(Z_1, Z_2, \dots, Z_w) \right). 
\end{equation}

\subsection{Optimization Strategy}
We adopt a unified objective by coupling the Kullback-Leibler (KL) regularization term with the Gaussian negative log-likelihood (NLL) term, jointly optimizing the multi-view sparse graph learning task and the sequence prediction task.

\paragraph{The short-term Branch}  
The Multi-View Sparse Graph Learner employs Gumbel-Softmax sampling to ultimately generate multi-level variable relationship graphs. The attention mask matrix \( \mathrm{probs}_{\mathrm{masked}} \) represents represent the edge probabilities for each head, and \( \pi \) denotes the prior distribution. As mentioned in Sec.~\ref{SARM}, to enhance the model's robustness and ensure the sparsity of the learned graphs, we introduce the prior probability \( \pi \) to guide the sparse graph learning process of each head. Specifically, we regularize each head with KL divergence to converge towards \( \pi \), ensuring the summation is performed only under the condition \( j \neq i \). Its overall computation can be formally formulated as:

\begin{equation}
\label{eq:kl}
L_{\mathrm{KL}}
=
\sum_{h=1}^{H}
\sum_{i=1}^{N}
\sum_{\substack{j=1\\ j\neq i}}^{N}
\mathrm{probs}^{(h)}_{\mathrm{masked},ij}
\log\!\left(
\frac{\mathrm{probs}^{(h)}_{\mathrm{masked},ij}}{\pi^{(h)}_{ij}}
\right).
\end{equation}

\paragraph{The Long-term Branch} 
The final predictions \(O\) are compared with ground truth \(Y\) under a Gaussian NLL, thereby improving the discrimination of subtle anomaly patterns:
\begin{equation}
\label{eq:nll}
L_{\mathrm{NLL}}
=
\frac{1}{2\sigma^2}
\sum_{w=1}^{W}
\sum_{n=1}^{N}
\sum_{d=1}^{d_{\mathrm{out}}}
\bigl(O_{w,n,d}-Y_{w,n,d}\bigr)^2
\;+\;
\frac{1}{2}\log\!\bigl(2\pi\sigma^2\bigr),
\end{equation}
where \(W\) denotes the target-window length, \(N\) refers to the number of variables, \(d_{\mathrm{out}}\) represents the output dimensionality, and \(\sigma^2\) denotes the variance parameter.

\paragraph{Overall Objective}
The overall training loss function combines structure regularization and likelihood fitting, which can be formally represented as:
\begin{equation}
\label{eq:total}
L
=
L_{\mathrm{KL}}
+
L_{\mathrm{NLL}}.
\end{equation}

\paragraph{Monitoring Signals}
During model training, we additionally track the channel-wise anomaly scores and a global score to quantify anomaly severity and support interpretability.

\section{Experiments}
\label{sec:exprs}

\subsection{Experimental Settings}
\subsubsection{Implementation Details}

Our experiments are implemented with the PyTorch framework and executed on a machine equipped with NVIDIA RTX 3090 GPUs. The model is optimized using Adam with L2 regularization and a learning-rate step-decay schedule. For the model configuration, we set the short-term window to $w=30$, the long-term history to $T=300$ with sliding step $s=5$. MSGL employs three heads of dimension 64 with prior $[0.9,0.05,0.05]$; DCGRU diffusion steps and multi-scale graph receptive fields are both 2; the bidirectional temporal-decay factor is dataset-dependent (0.7 / 0.8). All the models are trained for 200 epochs (batch 64, gradient-clip 1.0) with an initial learning rate of $10^{-3}$; the learning rate is reduced by a factor of 0.8 when the validation loss plateaus for 5 epochs (min LR $10^{-6}$); early stopping with a patience of 20 epochs is applied. To ensure the reproducibility, we run each experiment six times under fixed random seeds and report the averaged results.

\begin{table}[t]
  \centering
  \caption{Detailed characteristics of training and test set sizes, number of entities, monitored variables, and anomaly ratios on the MSL, SWaT, and WADI datasets.}
  \vspace{1em}
  \label{tab:datasets}
  \begin{tabular}{lccc}
    \toprule
    Datasets & MSL & SWaT & WADI \\
    \midrule
    Training Samples   & 58,317  & 496,800  & 1,045,200 \\
    Test Samples       & 73,729  & 449,919  & 172,800   \\
    Entities           & 27      & 1        & 1         \\
    Variables          & 55      & 51       & 118       \\
    Anomaly Ratio (\%) & 10.53   & 11.98    & 5.71      \\
    \bottomrule
  \end{tabular}
\end{table}

\subsubsection{Datasets}
Our experiments use three publicly available benchmark datasets. Table~\ref{tab:datasets} summarizes their key statistics. MSL~\cite{54} and SWaT\footnote{\url{https://itrust.sutd.edu.sg/itrust-labs-home/itrust-labs\_swat/}} each contain over 50 variables, whereas WADI\footnote{\url{https://itrust.sutd.edu.sg/itrust-labs-home/itrust-labs_wadi/}} exceeds 100. All three originate from operational technology (OT) systems or closely emulate OT environments. Additional details are provided in Appendix~A.1.

\subsubsection{Evaluation Metrics}

We evaluate anomaly detection performance using standard binary classification metrics: Precision, Recall, and F1. Moreover, we computed the average F1 score for each method across the three datasets. Additionally, to handle temporally continuous anomaly events, we adopt the point-adjusted evaluation strategy~\cite{53}, which considers a segment correctly detected if any point within it is identified as anomalous. All results reported in this paper follow this evaluation protocol.

\subsubsection{Baseline Methods}

We compare the proposed DARTs with several state-of-the-art methods in the field of Multivariate Time Series Anomaly Detection (MTSAD), including MAD-GAN~\cite{mad_gan}, OmniAnomaly~\cite{57}, MTAD-GAT~\cite{38}, USAD~\cite{20}, InterFusion~\cite{58}, TranAD~\cite{59}, ImDiffusion~\cite{9}, SimAD~\cite{62}, and CARLA~\cite{61}. Additional descriptions of the baselines are provided in Appendix A.2.

\begin{table*}[t!]
  \centering
  \caption{Performance comparison across three public datasets (dimension $>$ 50). 
  We report Precision ($P$), Recall ($R$), and F1 ($F1$) per dataset, and the average F1 ($Avg\_F1$) across datasets. 
  Best and second-best are in \textbf{bold} and \underline{underlined}, respectively.}
  \vspace{1em}
  \label{main_results}
  \renewcommand{\arraystretch}{1.22}
  \resizebox{\textwidth}{!}{%
  \begin{tabular}{l ccc ccc ccc c}
    \toprule
    \multirow{2}{*}{Method} &
      \multicolumn{3}{c}{WADI~(118d)} &
      \multicolumn{3}{c}{SWaT~(55d)} &
      \multicolumn{3}{c}{MSL~(51d)} &
      \multirow{2}{*}{\textbf{$Avg\_F1$~($\uparrow$)}} \\
    \cmidrule(lr){2-4}\cmidrule(lr){5-7}\cmidrule(lr){8-10}
    & $P$~($\uparrow$) & $R$~($\uparrow$) & $F1$~($\uparrow$)
    & $P$~($\uparrow$) & $R$~($\uparrow$) & $F1$~($\uparrow$)
    & $P$~($\uparrow$) & $R$~($\uparrow$) & $F1$~($\uparrow$) \\
    \midrule
    MAD\textendash GAN (ICANN 2019)         & 0.4698 & 0.2458 & 0.3200 & 0.7918 & 0.5423 & 0.6385 & 0.7047 & 0.7841 & 0.7423 & 0.5669 \\
    OmniAnomaly (SIGKDD 2019)               & 0.2811 & \underline{0.7051} & 0.4174 & 0.7223 & \textbf{0.9832} & 0.8328 & 0.9140 & 0.8891 & 0.8952 & 0.7151 \\
    MTAD\textendash GAT (ICDM 2020)         & 0.7792 & 0.5435 & 0.6404 & 0.9684 & 0.7866 & \underline{0.8681} & 0.8754 & 0.9440 & 0.9084 & \underline{0.8056} \\
    USAD (KDD 2020)                         & 0.6451 & 0.3220 & 0.4296 & \textbf{0.9870} & 0.7402 & 0.8460 & 0.8810 & \textbf{0.9786} & \underline{0.9109} & 0.7288 \\
    InterFusion (KDD 2021)                  & 0.5249 & 0.6552 & 0.5828 & 0.8683 & 0.8530 & 0.8600 & 0.7688 & \underline{0.9464} & 0.8442 & 0.7623 \\
    ImDiffusion (VLDB 2023)                 & 0.7029 & 0.5536 & 0.6194 & 0.8978 & \underline{0.8763} & \textbf{0.8868} & 0.8930 & 0.8638 & 0.8779 & 0.7947 \\
    SimAD (TNNLS 2025)                      & \underline{0.8125} & 0.5910 & \underline{0.6852} & 0.9581 & 0.7518 & 0.8425 & \underline{0.9198} & 0.8024 & 0.8571 & 0.7949 \\
    \textbf{DARTs (ours)}                   & \textbf{0.8939} & \textbf{0.7483} & \textbf{0.7110} & \underline{0.9706} & 0.7609 & 0.8531 & \textbf{0.9224} & 0.9115 & \textbf{0.9194} & \textbf{0.8278} \\
    \bottomrule
  \end{tabular}%
  }
  \vspace{2pt}
  {\footnotesize\emph{Notes:} $(\uparrow)$ indicates “higher is better”. Best/second-best marked by bold/underline.}
\end{table*}

\subsection{Main Results}

\subsubsection{Effectiveness and Robustness}
\label{Effectiveness and Robustness}
To validate the effectiveness of our DARTs design in high-dimensional data, we conducted extensive experiments on three publicly available benchmark datasets. As shown in Table~\ref{main_results}, the DARTs achieved state-of-the-art (SOTA) performance on the 118-dimensional WADI dataset, improving Precision, Recall, and F1 score by 10.02\%, 6.13\%, and 3.77\%, respectively, compared to the second-best method. These significant improvements highlight the superior performance and advantages of our approach in handling high-dimensional data.

Additionally, we also tested DARTs on the medium-dimensional SWaT and MSL datasets. The results demonstrate that our method exhibits strong competitive performance on medium-dimensional datasets as well. In particular, on the MSL dataset, DARTs achieved the best performance in both Precision and F1, surpassing the second-best method by 0.28\% and 0.93\%, respectively. On the SWaT dataset, although several baselines are specifically aligned with the dataset’s dimensional configuration, DARTs performed comparably to methods optimized for such datasets. This further confirms the generalizability and adaptability of our model, showing that it excels not only on high-dimensional data but also in lower-dimensional regimes.

Furthermore, considering the overall performance across the datasets, our method achieved the highest average F1 score across all three datasets, outperforming the second-best method by 2.76\%. This consistent superior performance emphasizes the broad applicability of DARTs across datasets of different sizes, and its robustness in handling noise from lower- to higher-dimensional data.

\begin{figure}[t]
  \centering
  \includegraphics[width=\columnwidth]{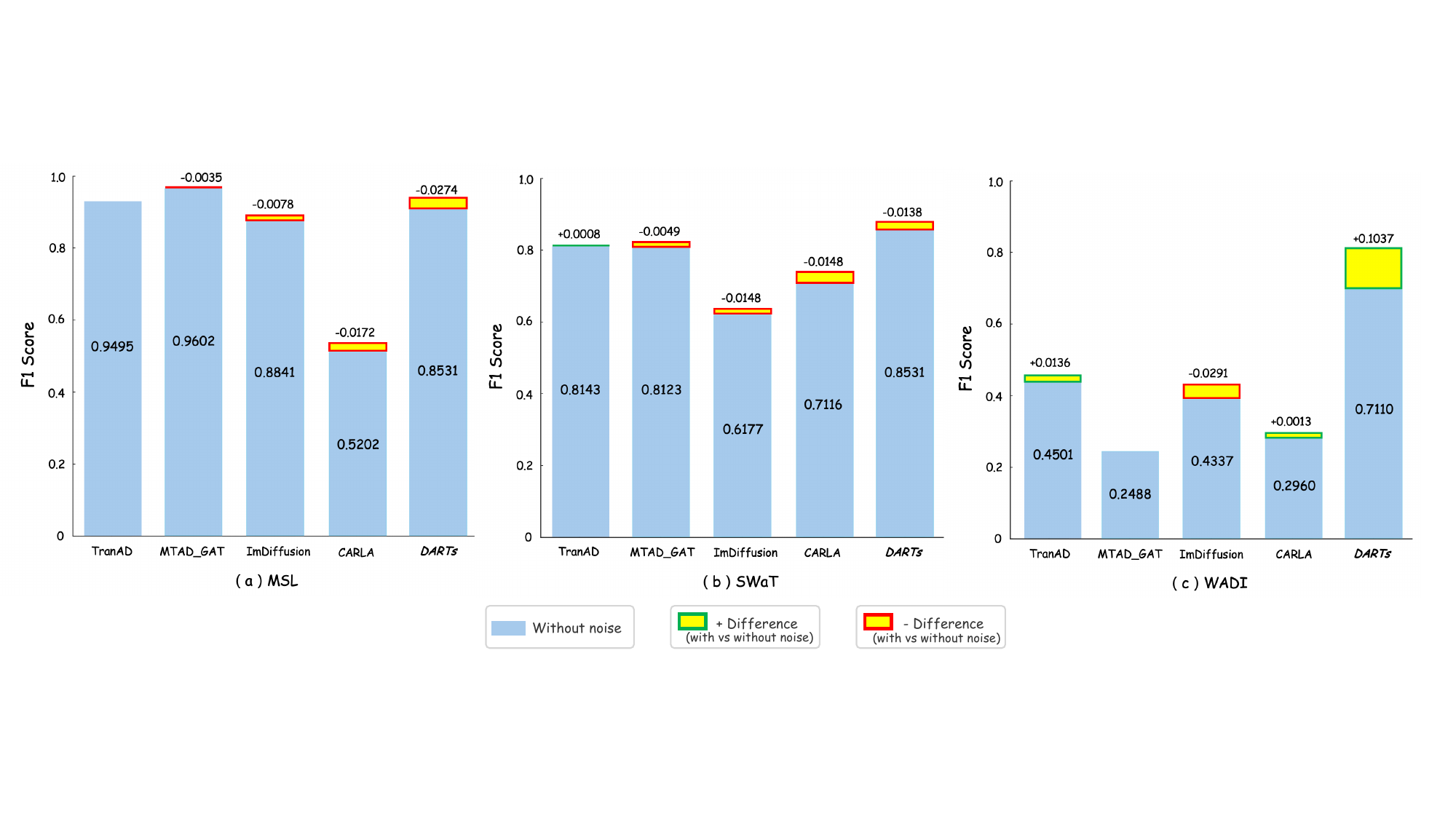}
  \vspace{-6.5em}
  \caption{Performance comparison of state-of-the-art (SOTA) methods and our proposed DARTs across three datasets, evaluated using both noisy and non-noisy training sets. The F1 score is used as the evaluation metric. Each of the three images corresponds to a different dataset, with bars representing the F1 scores under noisy conditions compared to non-noisy conditions. The baselines are evaluated using our data preprocessing settings.}
  \label{fig:robust}
\end{figure}

In addition, to further evaluate the robustness of DARTs in noisy environments, we introduced Gaussian noise (with a standard deviation of 0.5 times the standard deviation of each feature) to the training sets of all three datasets. As shown in Fig.~\ref{fig:robust}, we report the changes in F1 scores before and after adding noise for all methods. The results show that, in the noisy environment, DARTs achieved outstanding performance (0.8147) on the high-dimensional WADI dataset (Fig.~\ref{fig:robust}(c)), outperforming the second-best method (0.4637) by 75.7\%. Additionally, on the medium-dimensional SWaT dataset (Fig.~\ref{fig:robust}(b)), DARTs (0.8393) demonstrated a 2.97\% improvement over the second-best result (0.8151). On the MSL dataset (Fig.~\ref{fig:robust}(a)), DARTs also exhibited competitive performance, closely matching the second-best result. These results validate the exceptional robustness of DARTs in noisy environments, further confirming the effectiveness of our method in real-world complex settings.

It is important to note that the injection of noise did not lead to performance degradation for all methods across the datasets. In fact, our method, along with some others, exhibited improved performance on certain datasets after noise was introduced. This could be due to the fact that the noise helps the model generalize better and prevents overfitting, thereby enhancing its detection performance~\cite{robust}.

Overall, the experimental results presented above comprehensively demonstrate the outstanding performance of our method in high-dimensional data and its excellent robustness in noisy environments, further validating the effectiveness of the DARTs design and its promising potential for real-world applications.

\subsubsection{Intuition Validation}
\label{intuition validation}
We validate our intuitive hypothesis from both theoretical calculations and experimental verification:~Compared to a single-path architecture that processes both short-term windows and long-term history within the same branch, a dual-path architecture separates the processing of short-term windows and long-term history into two independent branches, demonstrating advantages in information oncentration capability and computational efficiency.

\paragraph{Theoretical}
The detailed procedure of our theoretical calculations for intuition validation is provided in Appendix B.

\begin{figure}[t]
  \centering
  \includegraphics[width=\linewidth]{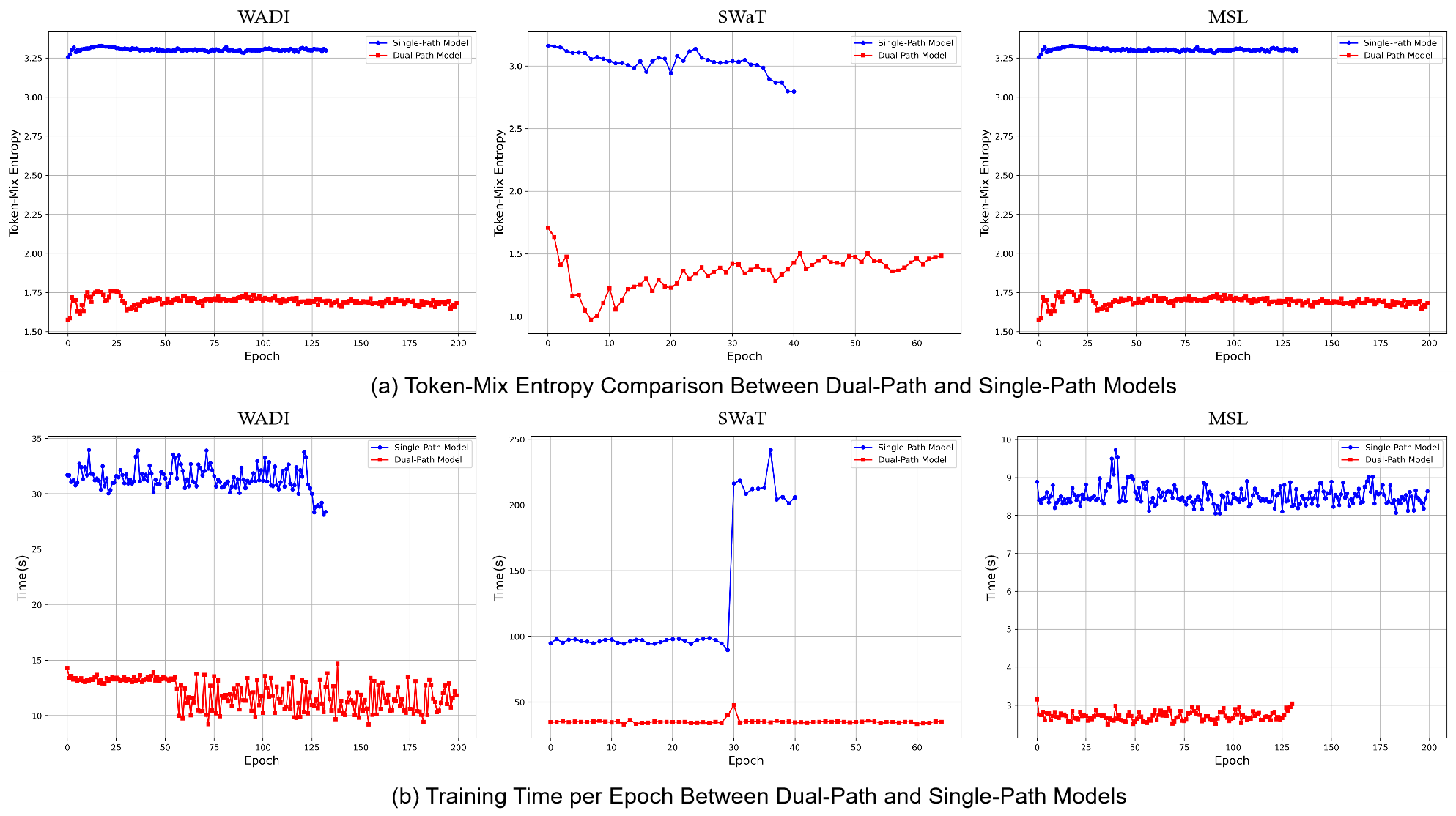}
  \caption{Comparison of (a) token-mix entropy, which evaluates information concentration, and (b) training time per epoch, which assesses model efficiency, between dual-path and single-path models across three datasets.}
  \label{fig:intution}
\end{figure}

\begin{table}[t!]
  \centering
  \caption{Comparison of peak memory usage and throughput across datasets.}
  \vspace{1em}
  \label{tab:performance_gains}
  \renewcommand{\arraystretch}{1.25}

  \resizebox{\linewidth}{!}{%
  \begin{tabular}{l *{3}{SS}}
    \toprule
    \multirow{2}{*}{Metrics} &
      \multicolumn{2}{c}{WADI (118d)} &
      \multicolumn{2}{c}{SWaT (55d)} &
      \multicolumn{2}{c}{MSL (51d)} \\
    \cmidrule(lr){2-3}\cmidrule(lr){4-5}\cmidrule(lr){6-7}
     & {Dual-Path} & {Single-Path} & {Dual-Path} & {Single-Path} & {Dual-Path} & {Single-Path} \\
    \midrule
    Peak Memory (MiB)~($\downarrow$)     & \bfseries 251.3 & 313.7 & \bfseries 261.6 & 339.7 & \bfseries 235.4 & 960.2 \\
    Throughput (samples/s)~($\uparrow$)  & \bfseries 64.3  & 28.8  & \bfseries 475.7 & 212.1 & \bfseries 356.4 & 54.7  \\
    \bottomrule
  \end{tabular}%
  }

  \vspace{2pt}
  {\footnotesize\emph{Notes:} $(\uparrow)$ higher is better; $(\downarrow)$ lower is better.}
\end{table}

\paragraph{Experimental}
We designed both single-path and dual-path models to experimentally validate our theoretical intuition. To ensure the validity of the experimental results, we selected the SARM branch as the core module of the model. Since the dual-path model requires a fusion module for dual-path outputs, whereas the single-path model can function without it, we introduced a simple linear fusion strategy to eliminate the potential impact of the fusion module, ensuring consistency between the two models in terms of fusion. This strategy is applied to both the single-path and dual-path models.

Specifically, in the single-path model, we retain only the SARM branch, using both the current window and the historical context as inputs. After processing, the output is split into two parts, in the same ratio of window to context as in the dual-path model. Each part is then mapped using a linear layer and fused by summation. In contrast, the dual-path model includes both the SARM and LSGM branches, with the window and context sequences processed independently by each branch. The outputs from both branches are then mapped with the same linear layers and fused by summation. This design ensures consistency in input data and fusion methods between the single-path and dual-path models, eliminating the influence of the fusion module and making the experimental results more comparable.

Based on the aforementioned design, we conducted experiments on three different datasets, testing token-mix entropy, training time per epoch (in seconds), Peak Memory (MiB), and Throughput (samples/s) during model training. Given the different scales of the datasets, some parameters differ slightly across datasets; however, the experimental configurations for the single-path and dual-path models remain consistent across all datasets, ensuring fairness in comparison.

The experimental results are presented in Fig.~\ref{fig:intution} and Tab.~\ref{tab:performance_gains}. Specifically, from Fig.~\ref{fig:intution} (a), we observe that the entropy of the single-path model is consistently higher than that of the dual-path model across all datasets, with a nearly doubling difference in all three datasets. This indicates that, in terms of information concentration, the single-path model, as predicted by our theoretical intuition, exhibits significantly lower information aggregation ability compared to the dual-path architecture.

In terms of computational efficiency, Fig.~\ref{fig:intution} (b) shows that the training time of the single-path model is consistently higher than that of the dual-path model. Notably, on the WADI and SWaT datasets, the training time of the single-path model is at least twice that of the dual-path model, while on the MSL dataset, the time consumption is nearly three times higher for the single-path model. This further validates the hypothesis that the computational efficiency of the single-path model is inferior to that of the dual-path model.

From Tab.~\ref{tab:performance_gains}, we see that Peak Memory reflects the maximum memory demand of the model during training. The results show that the dual-path model consistently requires less memory than the single-path model across all three datasets. Specifically, on the WADI and SWaT datasets, the maximum memory usage of the single-path model is approximately 1.25 to 1.3 times that of the dual-path model, and on the MSL dataset, the difference is even greater, approaching 4 times. This demonstrates that the dual-path model is more efficient in memory management, capable of performing computations with lower memory consumption.

Throughput, an important indicator of the model's ability to process samples per unit of time, further reflects computational efficiency. The results show that, on the WADI and SWaT datasets, the dual-path model processes more than twice the number of samples per second compared to the single-path model. On the MSL dataset, the difference is even more pronounced, with the dual-path model achieving a throughput that is six times higher than the single-path model. This result reinforces the significant advantage of the dual-path architecture in data processing efficiency.

In summary, the experimental results strongly support our theoretical hypothesis: in terms of information aggregation and computational efficiency, the dual-path model exhibits clear advantages over the single-path model, particularly in terms of throughput and memory management. These findings provide strong evidence for the effectiveness of the dual-path architecture and highlight its potential advantages in handling information at different temporal scales.

\subsubsection{Graph Learner Effectiveness}
\begin{figure}[t]
  \centering
  \includegraphics[width=0.75\linewidth]{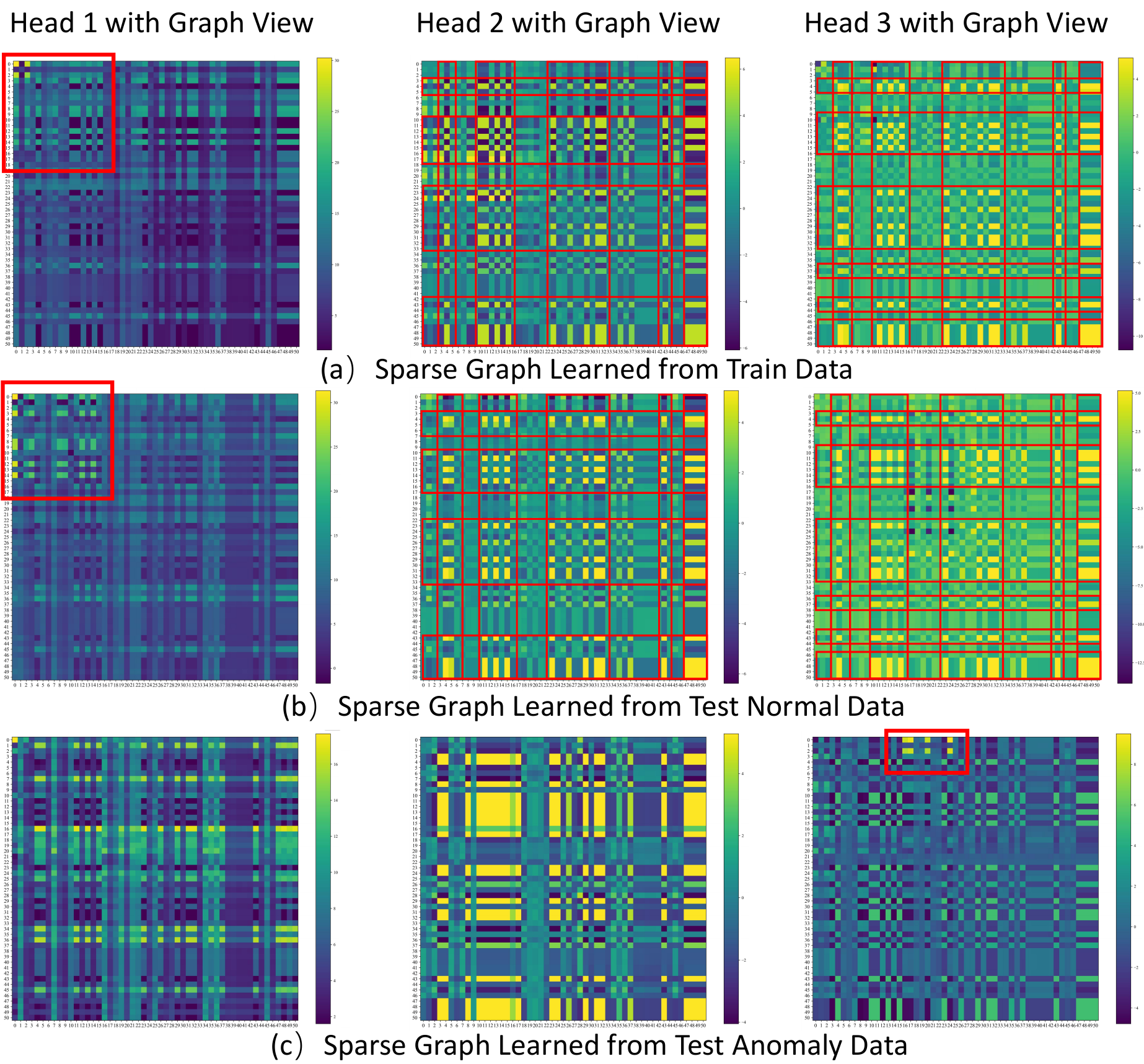}
  \caption{Three types of sparse graphs learned by the Multi-View Sparse Graph Learner on SWaT dataset: (a) training data, (b) test-time normal data, and (c) test-time anomaly data.}
  \label{fig:sparse_dependency_graphs}
\end{figure}

Fig.~\ref{fig:sparse_dependency_graphs} visualizes the multi-view sparse relation graphs adaptively learned by our DARTs, providing an insightful view of the inter-variable relationships captured by the system. Key structural patterns, critical for understanding the dynamics of the data, are highlighted with red boxes, emphasizing the most important connections and dependencies within the graph.

Upon examining the graphs learned from normal test segments (Fig.~\ref{fig:sparse_dependency_graphs}(b)), we observe that they remain consistent with those from the training phase (Fig.~\ref{fig:sparse_dependency_graphs}(a)). This consistency suggests that the model is able to reliably identify and preserve core inter-variable relation patterns, even in high-dimensional conditions. The model successfully captures the underlying structure of the data, enabling it to generalize effectively to unseen normal segments, thereby demonstrating robustness in handling complex, high-dimensional inputs.

In contrast, the graph derived from an anomalous segment (Fig.~\ref{fig:sparse_dependency_graphs}(c)) exhibits noticeable structural deviations. These deviations are marked by altered connectivity patterns, where the relationships between variables become disrupted, and abnormal hub nodes appear, signaling unusual variable interactions. Such anomalies suggest that the model is capable of detecting subtle changes in the underlying system behavior, even when those changes are not immediately obvious in the raw data. These structural anomalies serve as potential indicators of system faults or anomalies, which the model identifies by recognizing disruptions in the expected relational patterns. This capability underscores the model’s sensitivity and effectiveness in anomaly detection, particularly in high-dimensional, dynamic environments.

Furthermore, as mentioned in Sec.~\ref{SARM}, we guide the Multi-View Sparse Graph Learner using priors with significant disparities to ensure that it learns the most fundamental and simplistic spatiotemporal patterns, as well as more complex spatiotemporal patterns. Upon examining the graphs of normal sequences in Fig.~\ref{fig:sparse_dependency_graphs}(a) and Fig.~\ref{fig:sparse_dependency_graphs}(b), we observe that, under the experimental settings, head2 and head3 correspond to more intricate spatiotemporal patterns, while head1 represents the most basic spatiotemporal patterns.

\subsubsection{Anomaly Interpretation}
\label{Anomaly Interpretation}
\begin{figure}[t]
  \centering
  \includegraphics[width=0.8\linewidth]{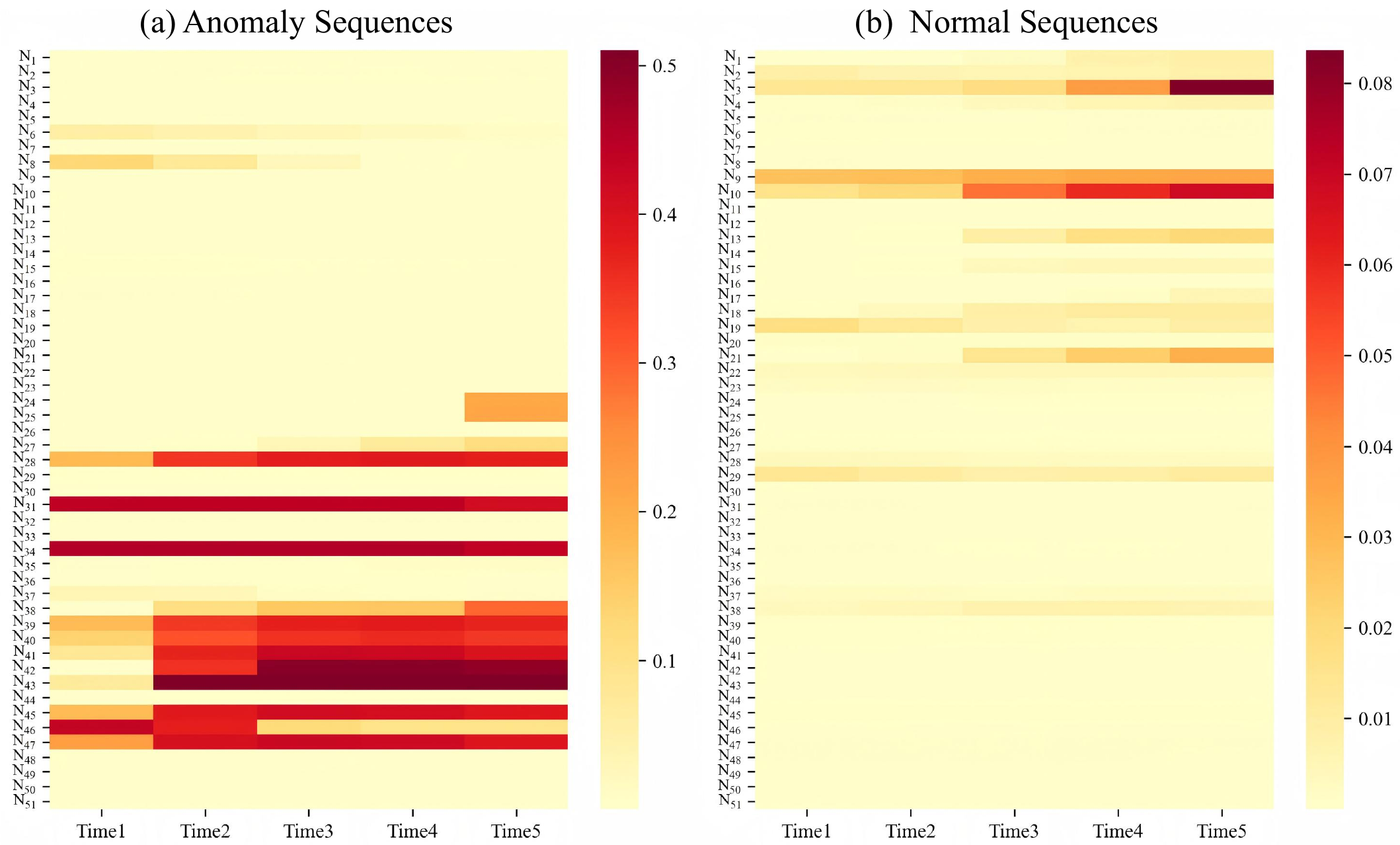}
  \caption{Interpretability of channel-wise anomaly scores over time on (a) anomalous and (b) normal sequences.}
  \label{fig:Interpretation}
\end{figure}

To validate the interpretability of the channel-wise anomaly scores computed by DARTs in relation to the corresponding physical sensor states, we visualized the channel-wise anomaly scores for anomalous sequences during the fifth attack (Fig.~\ref{fig:Interpretation}(a)) and normal sequences (Fig.~\ref{fig:Interpretation}(b)) on the SWaT dataset. It is important to note that the SWaT dataset contains six subsystems(Process 1-6), and the attack point in this case is variable 38 (sensor AIT-504), which belongs to the fifth subsystem (variables 35–47).

Overall, the channel-wise anomaly scores for the normal sequences consistently remain at low values, while the anomalous sequence exhibits a clear spatiotemporal anomaly propagation following the attack on variable 38 (sensor AIT-504): The anomaly gradually begins to emerge from Time2, peaks between Time3 and Time5, and spreads spatially from variable 38 (sensor AIT-504) to variables 39–47.

\begin{figure}[t]
  \centering
  \includegraphics[width=0.6\linewidth]{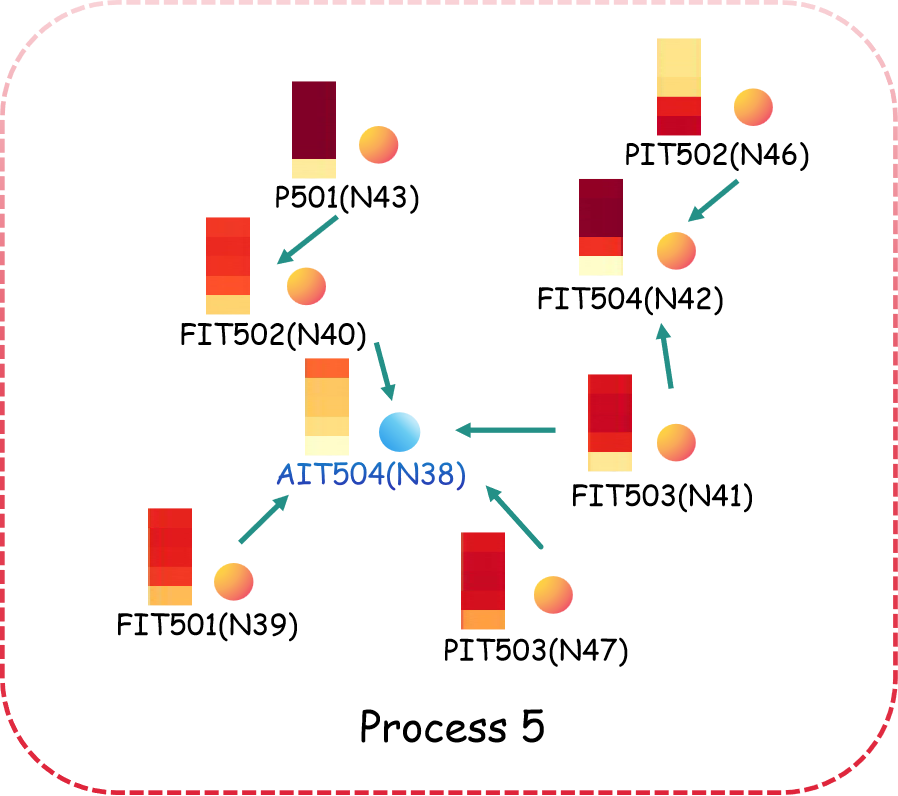}
  \caption{Core sensor interaction graph and its anomaly scores of Process 5.}
  \label{fig:Interpretation2}
\end{figure}

Fig.~\ref{fig:Interpretation2} further illustrates the interaction structure of the sensors corresponding to the group of anomalous variables affected by this anomaly propagation. During this attack, due to the system's compensation mechanisms, the anomaly score of the attack point, variable 38 (sensor AIT-504), remains relatively stable throughout the attack process. However, the anomaly scores of the multi-hop sensor group associated with it show an increase, indicating that DARTs is capable of detecting latent, indirect anomalies, even when the direct attack source remains unaffected. This finding highlights that DARTs can identify the group of anomalous variables in the spatiotemporal anomaly propagation based on channel anomaly scores, even though these anomalies are not immediately visible in the direct attack variable.

The interpretation of anomalies is not a strictly defined terminology. In recent methods, it is typically implemented in terms of root cause and severity level. The former refers to the features that contribute most to classifying a record as anomalous, while the latter refers to its duration~\cite{PR5}. In our experiment, the root cause is defined as the group of anomalous variables involved in the spatiotemporal anomaly propagation, which we distinguish based on the magnitude of the anomaly scores. The severity level, on the other hand, indicates the duration of the anomaly event. Moreover, the group of anomalous variables identified by DARTs includes indirect anomalous variables that are not directly impacted by the attack. This finding further suggests that DARTs is capable of accurately detecting anomalies caused by direct attacks as well as recognizing latent, indirect anomaly patterns, thereby enhancing the model's robustness and interpretability in complex attack scenarios.

\subsection{Ablation Studies}
\label{Ablation Studies}
To comprehensively evaluate the contribution of each architectural component in our DRATs, we conducted ablation experiments on the SWaT dataset. A detailed analysis of the ablation results is provided in Appendix C.1.

\subsection{Sensitivity Analysis}
 We perform a comprehensive sensitivity analysis on the SWaT dataset to evaluate the robustness of our proposed model across different hyperparameter configurations. A detailed analysis of the sensitivity results is provided in Appendix C.2.

\section{Conclusion and Discussions}
\label{sec:Conclusions}

In this work, we propose the DARTs model to address the challenge of multivariate time series anomaly detection in high-dimensional and noise-robust environments. First, based on intuitive conjectures, we designed a novel dual-path architecture to optimize the performance of traditional single-path models in high-dimensional data. Through both theoretical and experimental validation, we demonstrate that the dual-path model significantly outperforms the single-path model in terms of computational efficiency and information concentration capabilities. This result provides a more efficient and reliable solution for multivariate time series anomaly detection.

To further improve detection accuracy, we overcame the limitation of traditional models that rely solely on fixed time windows as input. We innovatively introduced historical context sequences as delayed anomaly signals for the instantaneous modes generated by short-term windows and designed a separate branch to model the spatiotemporal characteristics of this data. This design effectively enhanced the model's anomaly detection efficiency, and ablation studies further validated the effectiveness of this design, proving the necessity of incorporating historical context sequences.

Nevertheless, short-term windows remain a strongly correlated information source for anomaly detection compared to historical context. Therefore, in the short-term path, we designed a Multi-View Graph Learner, which successfully decouples the multi-level subsystems within the high-dimensional variables and constructs variable relationship graphs for each subsystem. By integrating the model’s channel anomaly scores, we can more accurately analyze the key anomalous variable groups within the anomaly windows, providing both theoretical support and practical pathways for the interpretability of anomaly detection. Our experiments on three medium- to high-dimensional datasets further confirm the advantages of this design.

With the continuous development of large-scale systems, we believe that future multivariate time series anomaly detection tasks should not be limited to low-dimensional public datasets, but should be designed to address more challenging industrial scenarios, particularly for anomaly detection in high-dimensional, redundant variable, and strong noise environments. Therefore, our dual-path architecture model, as a simple and efficient design, should become the preferred choice for this direction.

However, despite achieving promising results, hyperparameter selection remains a bottleneck in the application of the current model. Furthermore, the uneven spacing of the current dimensional settings limits our understanding of how performance improves with increasing dimensionality. To further enhance the model’s performance, future work will focus on constructing multi-level dimensional datasets to explore the model’s robustness and generalization across different dimensions, as well as implementing intelligent hyperparameter optimization. These future developments will advance the field of multivariate time series anomaly detection and ensure its widespread applicability and efficiency in real-world applications.

\bibliography{mybibfile}

\end{document}